%% file: root.tex
\documentclass{article}

\usepackage[final]{corl_2018} 

\input{preamble.tex}

\input{variables.tex}

\title{Learning What Information to Give in\\Partially Observed Domains}

\author{
  Rohan Chitnis \hspace{16mm} Leslie Pack Kaelbling \hspace{8mm} Tom\'as Lozano-P\'erez\\\\
  MIT Computer Science and Artificial Intelligence Laboratory\\
  \texttt{\{ronuchit, lpk, tlp\}@mit.edu}
}

\begin{document}
\maketitle


\begin{abstract}
  In many robotic applications, an autonomous agent must act within
  and explore a partially observed environment that is unobserved by
  its human teammate. We consider such a setting in which the agent
  can, while acting, transmit declarative \emph{information} to the
  human that helps them understand aspects of this unseen
  environment. In this work, we address the algorithmic question of
  how the agent should plan out what actions to take and what
  information to transmit. Naturally, one would expect the human to
  have \emph{preferences}, which we model information-theoretically by
  scoring transmitted information based on the change it induces in
  \emph{weighted entropy} of the human's belief state. We formulate
  this setting as a belief \mdp\ and give a tractable algorithm for
  solving it approximately. Then, we give an algorithm that allows the
  agent to learn the human's preferences online, through
  exploration. We validate our approach experimentally in simulated
  discrete and continuous partially observed search-and-recover
  domains. Visit \texttt{http://tinyurl.com/chitnis-corl-18} for a
  supplementary video.
\end{abstract}


\input{introduction.tex}
\input{related-work.tex}
\input{background.tex}
\input{problemsetting.tex}
\input{approach.tex}
\input{experiments.tex}
\input{conclusion.tex}

\clearpage
\acknowledgments{We gratefully acknowledge support from NSF grants
  1420316, 1523767, and 1723381; from AFOSR grant FA9550-17-1-0165;
  from Honda Research; and from Draper Laboratory. Rohan is supported
  by an NSF Graduate Research Fellowship. Any opinions, findings, and
  conclusions or recommendations expressed in this material are those
  of the authors and do not necessarily reflect the views of our
  sponsors.}


\bibliography{references}  

\end{document}

%% file: preamble.tex
\usepackage{multicol}
\usepackage{color}
\usepackage{amssymb}
\usepackage{amsmath}
\usepackage{amsthm}
\usepackage{tikz}
\usetikzlibrary{arrows,decorations.markings}

\usepackage{booktabs}
\usepackage{graphicx}
\usepackage{mathtools}
\usepackage{bbm}
\usepackage{wrapfig}
\usepackage{floatrow}
\newfloatcommand{capbtabbox}{table}[][\FBwidth]

\usepackage{caption}
\usepackage{subcaption}

\usepackage{algorithm2e}

\SetCommentSty{mycommfont}
\usepackage{indentfirst}

\DeclarePairedDelimiterX{\infdivx}[2]{(}{)}{%
  #1\;\delimsize\|\;#2%
}

\DeclareMathOperator*{\argmax}{argmax}

\newenvironment{tightlist}%
{\begin{list}{$\bullet$}{%
    \setlength{\topsep}{0in}
    \setlength{\partopsep}{0in}
    \setlength{\itemsep}{0in}
    \setlength{\parsep}{0in}
    \setlength{\leftmargin}{1.5em}
    \setlength{\rightmargin}{0in}
    \setlength{\itemindent}{-.1in}
}
}%
{\end{list}
}

\newcommand{\secref}[1]{Section~\ref{#1}}

\newcommand{\figref}[1]{Figure~\ref{#1}}
\newcommand{\algref}[1]{Algorithm~\ref{#1}}
\newcommand{\tabref}[1]{Table~\ref{#1}}

\newtheorem{thm}{Theorem}

\newtheorem{defn}{Definition}

%% file: variables.tex
\newcommand{\I}{\mathcal{I}}

\renewcommand{\P}{\mathcal{P}}
\newcommand{\St}{\mathcal{S}}
\newcommand{\A}{\mathcal{A}}

\newcommand{\Loss}{\mathcal{L}}

\newcommand{\B}{\mathcal{B}}
\newcommand{\D}{\mathcal{D}}
\newcommand{\pomdp}{{\sc pomdp}}
\newcommand{\mdp}{{\sc mdp}}
\renewcommand{\dag}{{\sc dag}}
\newcommand{\Obs}{\Omega}

%% file: introduction.tex
\section{Introduction}
\label{sec:introduction}
Consider a scenario where a human operator must manage several
autonomous search-and-rescue agents that can move through, observe,
and modify their respective environments, which are sites of recent
disasters. The agents have highly important objectives: to rescue
trapped victims. Secondarily, they should keep the human operator
informed about what is taking place, but should not sacrifice their
primary objective just to transmit information. Rather than receiving
a continuous stream of data such as a video feed from each agent, from
which it would be hard to extract salient findings, the human may only
want to receive important information, forcing the agents to make
decisions about what information is worth giving. Naturally, the human
will have \emph{preferences} about what information is important to
them: for instance, they would want to be notified when an agent
encounters a victim, but probably not every time it encounters a pile
of rubble.

In this work, we address the algorithmic question of how an agent
should plan out what actions to take in the world and what information
to transmit. We treat this problem as a sequential decision task where
on each timestep the agent can choose to transmit information, while
also acting in the world. To capture the notion that the human has
preferences, we model the human as an entity that scores the agent
based on how interesting the transmitted information is to them. The
agent's primary objective is to act optimally in the world;
secondarily, it should transmit score-maximizing information while
acting. We formulate this setting as a decomposable belief Markov
decision process (belief \mdp) and give a tractable algorithm for
solving it approximately in practice.

We model the human's score function information-theoretically. First,
we suppose that the human maintains a belief state, a probability
distribution over the set of possible environment states; this belief
gets updated based on information received from the agent. Next, we
let the human's score for a given piece of information be a function
of the change in \emph{weighted entropy} induced by the belief
update. This weighting is a crucial aspect of our approach: it allows
the human to describe, in a natural way, which aspects of the
environment they want to be informed about.

We give an algorithm that allows the agent to learn the human's
preferences online, through exploration. In this setting, online
learning is very important: the agent must explore in order to
discover the human's preferences, by giving them a variety of
information. We validate our approach experimentally in simulated
discrete and continuous partially observed search-and-recover domains,
and find that our belief \mdp\ framework and corresponding planning
and learning algorithms are effective in practice. Visit
\texttt{http://tinyurl.com/chitnis-corl-18} for a supplementary video.

%% file: related-work.tex
\section{Related Work}
\label{sec:relatedwork}
The problem setting we consider, in which an agent must act optimally
in its environment while secondarily giving information that optimizes
a human's score function, is novel but has connections to several
related problems in human-robot interaction. Our work is unique in
using weighted entropy to capture the human's preferences over which
aspects of the environment are important.

\emph{Information-theoretic perspective on belief updates.} The idea
of taking actions that lower the entropy of a belief state has been
studied in robotics for decades. Originally, it was applied to
navigation~\citep{actingunderuncertainty} and
localization~\citep{localizationentropy}. More recently, it has also
been used in human-robot interaction
settings~\citep{clarifyinfotheory, dialogueinfotheory}: the robot asks
the human clarifying questions about its environment to lower the
entropy of its own belief, which helps it plan more safely and
robustly. By contrast, in our method the robot is concerned with
estimating the entropy of the \emph{human's} belief, like in work by
\citet{roydialoguepomdp}.

\emph{Estimating the human's mental state.} Having a robot make
decisions based on its current estimate of the human's mental state
has been studied in human-robot collaborative
settings~\citep{theoryofmind, artcognition, actre}. The robot first
estimates the human's belief about the world state and goal, then uses
this information to build a human-aware policy for the collaborative
task. This strategy allows the robot to exhibit desirable behavior,
such as signaling its intentions in order to avoid surprising the
human.

\emph{Modeling user preferences with active learning.} The idea of
using active learning to understand user preferences has received
significant attention~\citep{activelearnhri, activelearnreward,
  pomdpelicit}. Typically in these methods, the agent gathers
information from the user through some channel, estimates a reward
function from this information, and acts based on this estimated
reward. Our method for learning the human's preferences online works
similarly, but we assume that the reward has an information-theoretic
structure.



%% file: background.tex
\section{Background}
\label{sec:background}
\subsection{Partially Observable Markov Decision Processes and Belief States}
\label{subsec:pomdpbackground}
Our work considers agent-environment interaction in the presence of
uncertainty, which is often formalized as a \emph{partially observable
  Markov decision process} (\pomdp)~\citep{pomdp}. An undiscounted
\pomdp\ is a tuple $\langle \St, \A, \Obs, T, O, R \rangle$: $\St$ is
the state space; $\A$ is the action space; $\Obs$ is the observation
space; $T(s, a, s') = P(s' \mid s, a)$ is the transition distribution
with $s, s' \in \St, a \in \A$; $O(s, o) = P(o \mid s)$ is the
observation model with $s \in \St, o \in \Obs$; and $R(s, a, s')$ is
the reward function with $s, s' \in \St, a \in \A$. Some states in
$\St$ are said to be \emph{terminal}, ending the episode and
generating no further reward. The agent's objective is to maximize its
overall expected reward,
$\mathbb{E}\left[\sum_{t}R(s_{t}, a_{t}, s_{t+1})\right]$. The optimal
solution to a \pomdp\ is a policy that maps the history of
observations and actions to the next action to take, such that this
objective is optimized. Exact solutions for interesting \pomdp s are
typically infeasible to compute, but some popular approximate
approaches are online
planning~\citep{pomcp,despot,beliefspaceplanning} and finding a policy
offline with a point-based solver~\citep{sarsop,pointpomdp}.

The sequence of states $s_{0}, s_{1}, ...$ is unobserved, so the agent
must instead maintain a \emph{belief state}, a probability
distribution over the space of possible states. This belief is updated
on each timestep, based on the received observation and taken
action. Unfortunately, representing this distribution exactly is
prohibitively expensive for even moderately-sized \pomdp s. One
typical alternative representation is a \emph{factored} one, in which
we assume the state can be decomposed into variables (features), each
with a value; the factored belief then maps each variable to a
distribution over possible values.

A \emph{Markov decision process} (\mdp)
$\langle \St, \A, T, R \rangle$ is a simplification of a \pomdp\ where
the states are fully observed by the agent, so $\Obs$ and $O$ are not
needed. The optimal solution to an \mdp\ is a policy that maps the
state to the next action to take, such that the same objective as
before is optimized.

Every \pomdp\ $\langle \St, \A, \Obs, T, O, R \rangle$ induces an
\mdp\ $\langle \B, \A, \tau, \rho \rangle$ on the belief space, known
as a \emph{belief \mdp}, where: $\B$ is the space of beliefs $B$ over
$\St$;
$\tau(B, a, B') = \sum_{o \in \Obs} P(B' \mid B, a, o) P(o \mid B,
a)$; and
$\rho(B, a, B') = \mathbb{E}_{s \sim B, s' \sim B'}R(s, a, s')$. See
\citet{pomdp} for details.

\subsection{Weighted Entropy and Weighted Information Gain}
\label{subsec:entropybackground}
Weighted entropy is a generalization of Shannon entropy that was first
presented and analyzed by \citet{weightedentropy}. The Shannon entropy
of a discrete probability distribution $p$, given by
$S(p) = \mathbb{E}\left[-\log p_{i}\right] = -\sum_{i:p_{i} \neq 0}
p_{i} \log p_{i}$, is a measure of the expected amount of information
carried by samples from the distribution, and can also be viewed as a
measure of the distribution's uncertainty. Note that trying to replace
the summation with integration for continuous distributions would not
be valid, because the interpretation of entropy as a measure of
uncertainty gets lost; e.g., the integral can be negative. The
information gain in going from a distribution $p$ to another $p'$ is
$S(p)-S(p')$.

\begin{defn}
  The \emph{weighted entropy} of a discrete probability distribution
  $p$ is given by
  $S_{w}(p) = -\sum_{i:p_{i} \neq 0} w_{i} p_{i} \log p_{i}$, where
  all $w_{i} \geq 0$. The \emph{weighted information gain} in going
  from a distribution $p$ to another $p'$ is $S_{w}(p)-S_{w}(p')$.
\end{defn}

Weighted entropy allows certain values of the distribution to
heuristically have more impact on the uncertainty, but cannot be
interpreted as the expected amount of information carried by samples.

\begin{wrapfigure}{l}{0.4\textwidth}
  \vspace{-1em}
  \centering
    \noindent
    \includegraphics[scale=0.35]{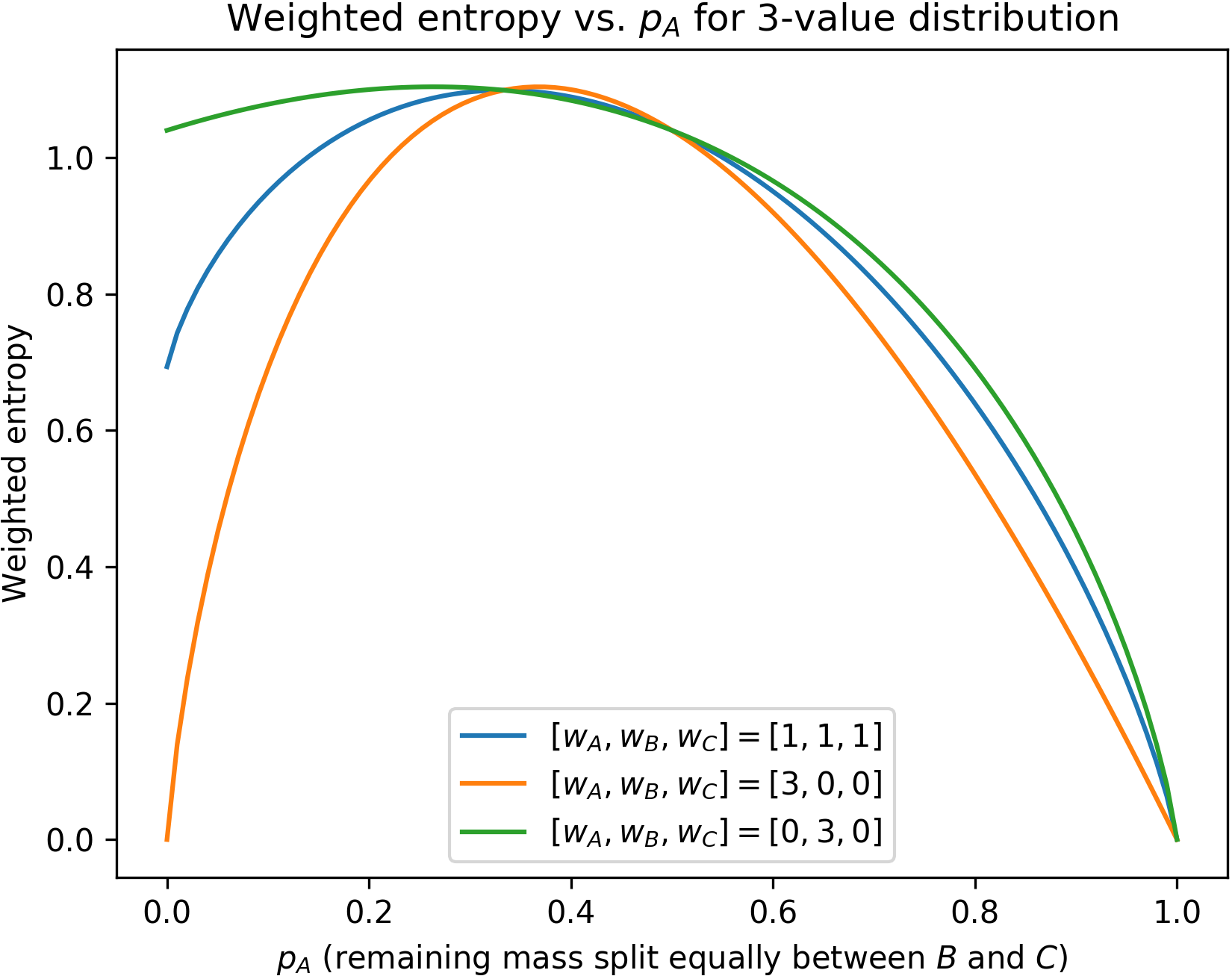}
    \caption{\small{Weighted entropy for a distribution with three
        values: $A, B, C$. The x-axis varies $p_{A}$, with the
        remaining probability mass split equally between $B$ and
        $C$.}}
  \label{fig:weightedentropy}
\end{wrapfigure}

\emph{Intuition.} \figref{fig:weightedentropy} helps give intuition
about weighted entropy by plotting it for the case of a distribution
with three values. In the figure, we only let $p_{A}$ vary freely and
set $p_{B} = p_{C} = \frac{1-p_{A}}{2}$, so that the plot can be
visualized in two dimensions. When only one value is possible
($p_{A} = 1$), the entropy is always 0 regardless of the setting of
weights, but as $p_{A}$ approaches 1 from the left, the entropy drops
off more quickly the higher $w_{A}$ is (relative to $w_{B}$ and
$w_{C}$). If all weight is placed on $A$ (the orange curve), then when
$p_{A} = 0$ the entropy also goes to 0, because the setting of weights
conveys that distinguishing between $B$ and $C$ gives no
information. However, if no weight is placed on $A$ (the green curve),
then when $p_{A} = 0$ we have $p_{B} = p_{C} = 0.5$, and the entropy
is high because the setting of weights conveys that all of the
information lies in telling $B$ and $C$ apart.

%% file: problemsetting.tex
\section{Problem Setting}
\label{sec:formulation}
\begin{figure}[h]
  \centering
    \noindent
    \includegraphics[width=0.8\textwidth]{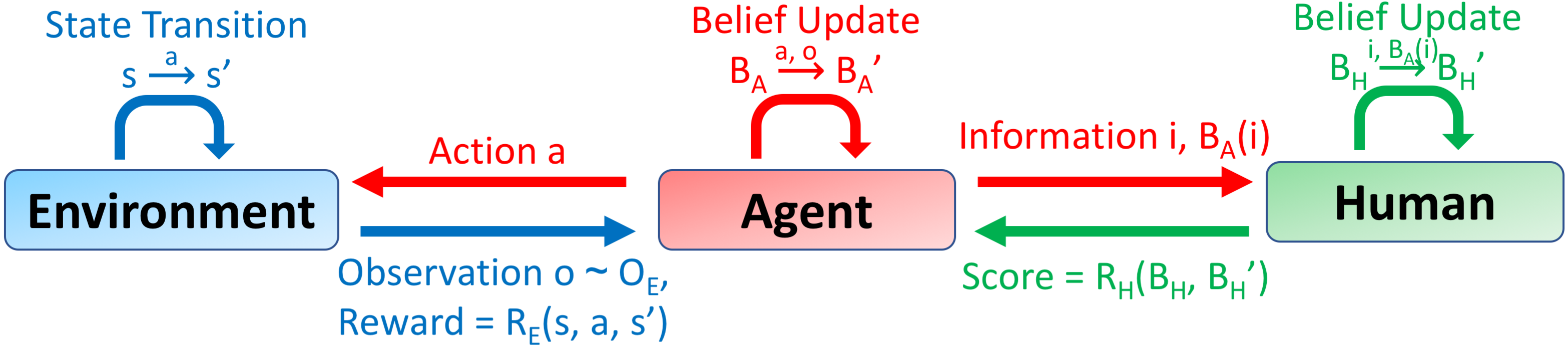}
    \caption{\small{A diagram of our problem setting. Red: agent's
        activities; blue: environment's; green: human's.}}
  \label{fig:problemsetting}
\end{figure}

We formulate our problem setting as a belief \mdp\
(\secref{subsec:pomdpbackground}) from the agent's perspective, then
give an algorithm for solving it approximately. At each timestep, the
agent takes an action in the environment and chooses a piece of
information $i$ (or null if it chooses not to give any) to transmit,
along with the marginal probability, $B_{A}(i)$, of $i$ under the
agent's current belief. See \figref{fig:problemsetting}.

Our presentation of the formulation will assume that the agent knows
1) the human's initial belief, 2) the model for how the human updates
their belief, and 3) that only information from the agent can induce
belief updates; this assumption effectively renders the human's belief
state fully observed by the agent. We can easily relax this
assumption: suppose the agent were allowed to query only some aspects
of the human's belief; then, it could incorporate the remainder into
its own belief as part of the latent state. We will not complicate our
presentation by describing this setting explicitly.

\subsection{Belief \mdp\ Formulation}
Let the agent-environment interaction be modeled as a \pomdp\
$\langle \St_{E}, \A_{E}, \Obs_{E}, T_{E}, O_{E}, R_{E} \rangle$,
where $\St_{E}$ is continuous or discrete. This induces a belief \mdp\
$\langle \B_{E}, \A_{E}, \tau_{E}, \rho_{E} \rangle$, where $\B_{E}$
is the space of beliefs over $\St_{E}$. The agent maintains a belief
state $B_{A} \in \B_{E}$, updated with a Bayes filter~\citep{aima}.

The human maintains their own belief state $B_{H} \in \B_{E}$ over
environment states, updated based only on information transmitted by
the agent, and gives the agent a real-valued score on each timestep
for this information. We model the human as a tuple
$\langle \I, T_{H}, B_{H}^{0}, R_{H} \rangle$: $\I$ is a set of
fluents (Boolean atoms that may or may not hold in the state) that
defines the space of information the agent can transmit;
$T_{H}(s, s') = P(s' \mid s)$ is the human's forward model of the
world with $s, s' \in \St_{E}$; $B_{H}^{0} \in \B_{E}$ is the human's
initial belief; and $R_{H}(B_{H}, B_{H}')$ is the human's score
function with $B_{H}, B_{H}' \in \B_{E}$. The $T_{H}$ allows the human
to model the degradation of information over time; we use a simple
$T_{H}$ that is almost the identity function, but gives $\epsilon$
probability to non-identity transitions.

At each timestep, the agent selects information $i \in \I$ to give and
transmits it along with the marginal probability of $i$ under $B_{A}$,
defined as
$B_{A}(i) = \sum_{s \in \St_{E}: i\text{ holds in }s} B_{A}(s)$. We
update the belief $B_{H}$ according to Jeffrey's
rule~\citep{jeffreysrule}, which is based on the principle of
\emph{probability kinematics} for minimizing the change in
belief. First, we define
$\tilde{B}_{H}(s) = \sum_{s' \in \St_{E}} T_{H}(s', s)B_{H}(s'), \forall
s \in \St_{E}$. Then the full belief update, $B_{H} \to B_{H}'$, is
$B_{H}'(s) = \frac{\tilde{B}_{H}(s)B_{A}(i)}{\tilde{B}_{H}(i)}$ if $i$
holds in $s$ and $\frac{\tilde{B}_{H}(s)(1-B_{A}(i))}{1-\tilde{B}_{H}(i)}$
if $i$ does not hold in $s$, $\forall s \in \St_{E}$. The summations
can be replaced with integration if $\St_{E}$ is continuous.

\emph{Objective.} We define the agent's objective as follows: to act
optimally in the environment (maximizing the expected sum of rewards
$R_{E}$) and, subject to acting optimally, to give information such
that the expected sum of the human's scores $R_{H}$ over the
trajectory is maximized.

The full belief \mdp\ $\P$ for this setting (from the agent's
perspective) is a tuple $\langle \B, \A, \tau, \rho \rangle$:
\begin{tightlist}
\item $\B = \B_{E} \times \B_{E}$. A state is a pair of the agent's
  belief $B_{A} \in \B_{E}$ and the human's belief $B_{H} \in \B_{E}$.
\item $\A = \A_{E} \times \I$. An action is a pair of environment
  action $a \in \A_{E}$ and information $i \in \I$.
\item
  $\tau(\langle B_{A}, B_{H} \rangle, \langle a, i \rangle, \langle
  B_{A}', B_{H}' \rangle) = \tau_{E}(B_{A}, a, B_{A}')$ if $B_{H}'$
  satisfies the update equation, else 0.
\item
  $\rho(\langle B_{A}, B_{H} \rangle, \langle a, i \rangle, \langle
  B_{A}', B_{H}' \rangle)$ is a pair
  $\langle \rho_{E}(B_{A}, a, B_{A}'), R_{H}(B_{H}, B_{H}') \rangle$
  with the comparison operation
  $\langle x_{1}, y_{1} \rangle > \langle x_{2}, y_{2} \rangle \iff
  x_{1} > x_{2} \lor (x_{1} = x_{2} \land y_{1} > y_{2})$; similarly
  for $<$.
\end{tightlist}

The following algorithm for solving $\P$ by decomposition will help us
give an approximation next.

\setlength{\textfloatsep}{10pt}
\begin{algorithm}[h]
\begin{small}
  \SetAlgoLined
  \SetAlgoNoEnd
  \DontPrintSemicolon
  \SetKwFunction{algo}{algo}\SetKwFunction{proc}{proc}
  \SetKwProg{myalg}{Algorithm}{}{}
  \SetKwProg{myproc}{Subroutine}{}{}
  \SetKw{Continue}{continue}
  \SetKw{Return}{return}
  \SetKw{Emit}{emit}
  \myalg{\textsc{DecomposeAndSolve}($\P$)}{
    \nl $\pi_{\text{act}} \gets$ (solve agent-environment belief \mdp\ $\langle \B_{E}, \A_{E}, \tau_{E}, \rho_{E} \rangle$)\;
    \nl \tcp{Define $\tau_{H}$ as $\tau_{H}(\langle B_{A}, B_{H} \rangle, i, \langle B_{A}', B_{H}' \rangle) = \tau(\langle B_{A}, B_{H} \rangle, \langle \pi_{\text{act}}(B_{A}), i \rangle, \langle B_{A}', B_{H}' \rangle)$.}
    \nl \tcp{Define $\rho_{H}$ as $\rho_{H}(\langle B_{A}, B_{H} \rangle, i, \langle B_{A}', B_{H}' \rangle) = R_{H}(B_{H}, B_{H}')$.}
    \nl $\pi_{\text{info}} \gets$ (solve agent-human belief \mdp\ $\langle \B_{E} \times \B_{E}, \I, \tau_{H}, \rho_{H} \rangle$)\;
    \nl \Return policy $\pi$ for $\P$: $\pi(\langle B_{A}, B_{H} \rangle) = \langle \pi_{\text{act}}(B_{A}), \pi_{\text{info}}(\langle B_{A}, B_{H} \rangle) \rangle$\;
    }\;
\end{small}
\caption{\small{Algorithm for solving $\P$ by decomposition. The
    agent-human belief \mdp\ must include the agent's belief $B_{A}$
    in the state so that the marginal probabilities of information,
    $B_{A}(i)$, can be determined.}}
\label{alg:optplanning}
\end{algorithm}

\begin{thm}
  \label{thm1}
  \algref{alg:optplanning} returns an optimal solution $\pi^{*}$ for $\P$.
\end{thm}

\emph{Proof.} Note that a policy $\pi$ for $\P$ maps pairs
$\langle B_{A}, B_{H} \rangle$ to pairs $\langle a, i \rangle$, with
$a \in \A_{E}$ and $i \in \I$. We have
$\pi^{*} = \argmax_{\pi} \mathbb{E}\left[\sum_{t}\rho(\langle B_{A,t},
  B_{H,t} \rangle, \pi(\langle B_{A,t}, B_{H,t} \rangle), \langle
  B_{A,t+1}, B_{H,t+1} \rangle)\right]$. Define
$\pi(\langle B_{A}, B_{H} \rangle)[0] = a$, the first entry in the
pair. Due to the comparison operation we defined on $\rho$, we can
write
$\pi^{*} = \argmax_{\pi} \mathbb{E}\left[\sum_{t}\rho_{E}(B_{A,t},
  \pi(\langle B_{A,t}, B_{H,t} \rangle)[0], B_{A,t+1})\right]$, and if
there are multiple such $\pi^{*}$, pick the one that also maximizes
$\mathbb{E}\left[\sum_{t}R_{H}(B_{H,t}, B_{H,t+1})\right]$. The
decomposition strategy exactly achieves this, by leveraging the fact
that the human cannot affect the environment. \qed

\subsection{Approximation Algorithm}
$\P$ can be hard to solve optimally even using the decomposition
strategy of \algref{alg:optplanning}. A key challenge is that
$\pi_{\text{act}}$ branches due to uncertainty about observations and
transitions, so searching for the optimal $\pi_{\text{info}}$ becomes
computationally infeasible. Instead, we apply the
determinize-and-replan strategy~\citep{plattmlobs,bsptamp,ffreplan},
which is not optimal but often works well in practice. We determinize
$\P$ using a maximum likelihood assumption~\citep{plattmlobs}, then
use \algref{alg:optplanning}. This procedure is repeated any time the
determinization is found to have been violated. See
\algref{alg:planning} for full pseudocode.

Line 3 generates the trajectory $\tau_{B_{A}}$ of the agent's beliefs
induced by $p_{\text{act}}$, which works because $p_{\text{act}}$ does
not contain branches. Line 8 constructs a directed acyclic graph
(\dag) $G$ whose states are tuples of (human belief, timestep). An
edge exists between $(B_{H}, t)$ and $(B_{H}', t+1)$ iff some
information $i \in \I$ causes the belief update $B_{H} \to B_{H}'$
under the determinized $\P$.  The edge weight is
$R_{H}(B_{H}, B_{H}')$, the human's score for $i$. Note that all paths
through $G$ have the same number of steps, and because the edge
weights are the human's scores, the longest weighted path through $G$
is precisely the information-giving plan $p_{\text{info}}$ that
maximizes the total score over the trajectory. Our implementation does
not build the full \dag\ $G$; we prune the search using
domain-specific heuristics.

\setlength{\textfloatsep}{10pt}
\begin{algorithm}[h]
\begin{small}
  \SetAlgoLined
  \SetAlgoNoEnd
  \DontPrintSemicolon
  \SetKwFunction{algo}{algo}\SetKwFunction{proc}{proc}
  \SetKwProg{myalg}{Algorithm}{}{}
  \SetKwProg{myproc}{Subroutine}{}{}
  \SetKw{Continue}{continue}
  \SetKw{Return}{return}
  \SetKw{Emit}{emit}
  \myalg{\textsc{DecomposeAndSolveApproximate}($\P$)}{
    \nl $\P$.Determinize()\;
    \nl $p_{\text{act}} \gets$ (solve agent-environment portion of $\P$) \tcp*{\footnotesize Acting plan (no branches).}
    \nl $\tau_{B_{A}} \gets$ (trajectory of beliefs $B_{A}$ induced by $p_{\text{act}}$)\;
    \myproc{\textsc{GetSuccessors}(state)}{
      \nl $(B_{H}, \text{timestep}) \gets$ state \tcp*{\footnotesize Unpack state tuple.}
      \nl \For{each $i \in \I$}{
        \nl $B_{H}' \gets$ (result of updating $B_{H}$ with $i$ and marginal probability $\tau_{B_{A}}[\text{timestep}](i)$)\;
        \nl \Emit next state $(B_{H}', \text{timestep}+1)$ with edge label $i$ and weight $R_{H}(B_{H}, B_{H}')$\;
      }
    }
    \nl $G \gets$ (\dag\ constructed from root node $(B_{H}^{0}, 0)$ and \textsc{GetSuccessors})\;
    \nl $p_{\text{info}} \gets$ \textsc{LongestWeightedPathDAG}($G$) \tcp*{\footnotesize Information-giving plan (no branches).}
    \nl \Return \textsc{Merge}($p_{\text{act}}, p_{\text{info}}$) \tcp*{\footnotesize Zip into a single plan.}
    }\;
\end{small}
\caption{\small{Approximate approach for solving $\P$ with
    determinization. See text for detailed description.}}
\label{alg:planning}
\end{algorithm}

%% file: approach.tex
\section{Learning an Information-Theoretic Score Function}
\label{sec:approach}
In this section, we first model the human's score function $R_{H}$
information-theoretically using the notion of weighted entropy. Then,
we give an algorithm by which the agent can learn $R_{H}$ online.

\subsection{Score Function Model}
\label{subsec:model}
We model the score function $R_{H}(B_{H}, B_{H}')$ as a function
$f \in \mathbb{R}$ of the weighted information gain
(\secref{subsec:entropybackground}) of the belief update induced by
information:
$$R_{H}(B_{H}, B_{H}') = f(S_{w}(B_{H})-S_{w}(B_{H}')),$$ where the $w$
are a set of weights. The human chooses both $w$ and $f$ to suit their
preferences.

\emph{Assumptions.} This model introduces two assumptions. 1) The
human's belief $B_{H}$, which is ideally over the environment state
space $\St_{E}$, must be over a discrete space in order for its
entropy to be well-defined. If $\St_{E}$ is continuous, the human can
make any discrete abstraction of $\St_{E}$, and maintain $B_{H}$ over
this abstraction instead of over $\St_{E}$. Note that the agent must
know this discrete abstraction. 2) If the belief is factored
(\secref{subsec:pomdpbackground}), we calculate the total entropy by
summing the entropy of each factored distribution. This is an upper
bound that assumes independence among the factors.

\emph{Motivation.} Assuming structure in the form of $R_{H}$ makes it
easier for the agent to learn the human's preferences; the notion of
weighted entropy is a compelling choice. The human's belief state
$B_{H}$ captures their perceived likelihood of each possible
environment state (or value of each factor in the state). Each $p_{i}$
term in the entropy formula corresponds to an environment state or
value of a factor, so the $w_{i}$ encode the human's preferences about
which states or values of factors are important.

\emph{Interpretation of $f$.} Different choices of $f$ allow the human
to exhibit various preferences. Choosing $f$ to be the identity
function means that the human wants the agent to act greedily,
transmitting the highest-scoring piece of information at each
timestep. The human may instead prefer for $f$ to impose a threshold
$t$: if the gain is smaller than $t$, then $f$ could return a negative
score to penalize the agent for not being sufficiently informative. A
sublinear $f$ rewards the agent for splitting up information into
subparts and transmitting it over multiple timesteps, while a
superlinear $f$ rewards the agent for withholding partial information
in favor of occasionally transmitted, more complete information.

\subsection{Learning Preferences Online}
\label{subsec:approachlearn}
We now give \algref{alg:learning}, which allows the agent to learn $w$
and $f$ online through exploration. This algorithm works for both
single-episode lifelong learning problems where no states are terminal
and short-horizon problems where the agent must learn over several
episodes. In Line 7, the agent explores the human's preferences using
an $\epsilon$-greedy policy that gives a random piece of information
with probability $\epsilon$ and otherwise follows $\pi$, the policy
solving $\P$ under the current $\hat{w}$ and $\hat{f}$.

If the human's preferences ($w$ or $f$) ever change, we can reset
$\epsilon$ to an appropriate value and continue running the algorithm,
so the agent can explore information that the human now finds
interesting. Additionally, with some small modifications we can make
$f$ depend on the last few timesteps of transmitted information: we
need only augment states in our belief \mdp\ with this history so it
can be used to calculate $R_{H}$, and include this history in the
dataset $\D$ used for learning in \algref{alg:learning}.

\setlength{\textfloatsep}{10pt}
\begin{algorithm}[h]
\begin{small}
  \SetAlgoLined
  \SetAlgoNoEnd
  \DontPrintSemicolon
  \SetKwFunction{algo}{algo}\SetKwFunction{proc}{proc}
  \SetKwProg{myalg}{Algorithm}{}{}
  \SetKwProg{myproc}{Subroutine}{}{}
  \SetKw{Continue}{continue}
  \SetKw{Return}{return}
  \SetKw{Emit}{emit}
  \myalg{\textsc{TrainLoop}}{
    \nl $\hat{w}, \hat{f} \gets$ (initial guess)\;
    \nl $\D \gets$ (initialize empty dataset)\;
    \nl $\P \gets$ (initialize problem) \tcp*{\footnotesize Belief \mdp\ described in \secref{sec:formulation}.}
    \nl $\pi \gets$ \textsc{Solve}($\P, \hat{w}, \hat{f}$) \tcp*{\footnotesize Solve $\P$ under $\hat{w}$ and $\hat{f}$ (e.g., with Alg.~\ref{alg:planning}).}
    \nl \While{not done training}{
        \nl Act according to $\pi$ in environment.\;
        \nl Give information according to $\epsilon$-greedy$\infdivx{\pi}{\text{random}}$; obtain noisy human's score $\tilde{s}$.\;
        \nl Store tuple of transition and noisy score, $(B_{H}, B_{H}', \tilde{s})$, into $\D$.\;
        \nl Sample training batch $T \sim \D$.\;
        \nl $\Loss \gets \sum_{T}(\tilde{s}-\hat{f}(S_{\hat{w}}(B_{H})-S_{\hat{w}}(B_{H}')))^{2}/T.\text{size}$ \tcp*{\footnotesize Loss is {\sc mse} of predicted score.}
        \nl Update $\hat{w}, \hat{f}$ with optimization step on $\Loss$.\;
        \nl \If{agent reaches terminal state}{
          \nl Repeat Lines 3-4.\;
        }
      }}\;
\end{small}
\caption{\small{Training loop for estimating the human's true $w$ and $f$,
    given noisy supervision.}}
\label{alg:learning}
\end{algorithm}

%% file: experiments.tex
\section{Experiments}
\label{sec:experiments}
We show results for three settings of the function $f$: identity,
square, and natural logarithm. All three use a threshold $t = 1$: if
the weighted information gain is less than 1, then $f$ returns $-10$,
penalizing the agent. (This threshold is arbitrary, as the weights can
always be rescaled to accommodate any threshold.) If the information is
null, then $f$ returns $10^{-3}$, which causes the agent to slightly
prefer giving no information rather than information that induces no
change in the human's belief. We use the same weights $w$ for each
factor in the belief, though this simplification is not required.

We implemented $\hat{w}$ and $\hat{f}$ in
TensorFlow~\citep{tensorflow} as a single fully connected network,
with hidden layer sizes [100, 50], that outputs the predicted
score. The model takes as input a vector of the change, between
$B_{H}$ and $B_{H}'$, in $p_{i} \log p_{i}$ for each entry $p_{i}$ in
the belief. We used a gradient descent optimizer with learning rate
$10^{-1}$, $\ell_{2}$ regularization scale $10^{-7}$, sigmoid
activations, batch size 100, and $\epsilon$ exponentially decaying
from 1 to roughly $10^{-2}$ over the first 20 episodes.

We experiment with simulated discrete and continuous partially
observed search-and-recover domains, where the agent must find and
recover objects in the environment while transmitting information
about these objects based on the human's preferences. Although the
\pomdp s we consider are simple, the aim of our experiments is to
understand and analyze the nature of the transmitted information, not
to require the agent to plan out long sequences of actions in the
environment.

\begin{figure}[t]
  \centering
    \noindent
    \includegraphics[width=\textwidth]{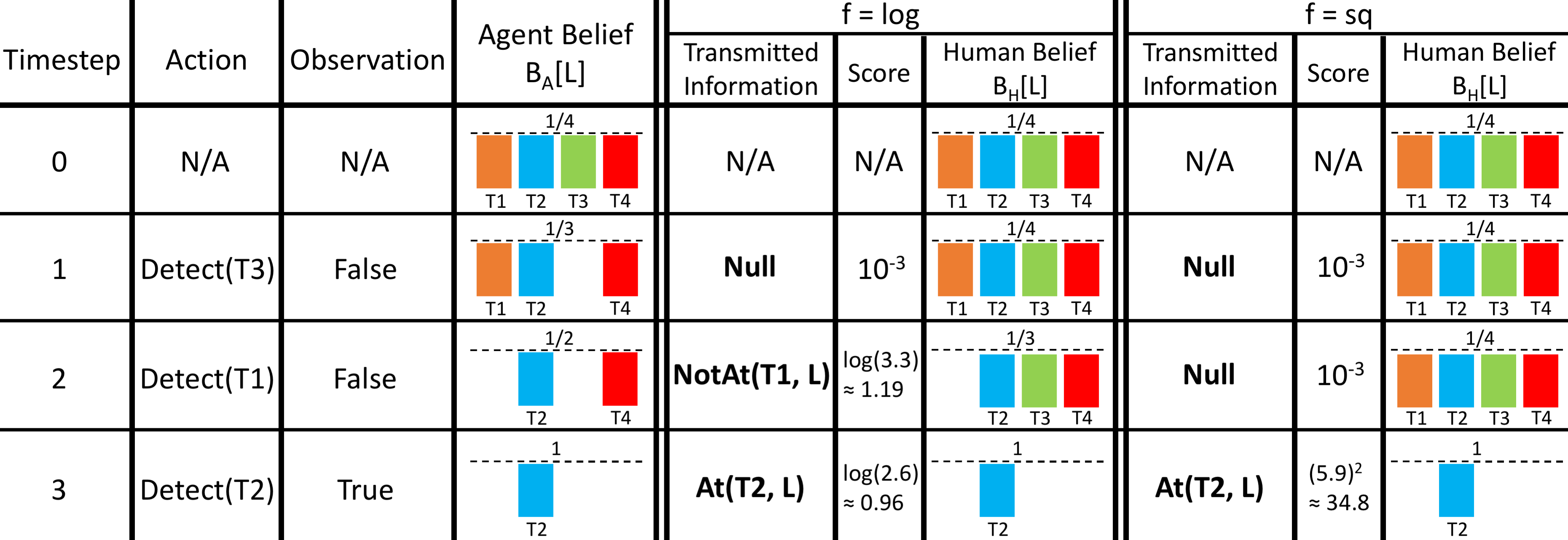}
    \caption{\small{Example execution of Domain 1 with a single
        location L, showing how $w$ and $f$ affect the optimal
        transmitted information. For this example, the agent knows
        $R_{H}$ (no learning). T1, T2, T3, and T4 are the object
        types. At each timestep, the agent \textsc{detect}s whether
        the object at L is of a particular type, and updates its
        belief $B_{A}$[L] accordingly. The human's weights $w$ are
        \{T1: 10, T2: 5, T3: 1, T4: 1\}, and $f$ uses a threshold
        $t = 1$ as discussed in the text. Agent and human beliefs are
        initialized uniformly over object types. Key points: (1) the
        agent chooses not to transmit the information NotAt(T3, L) in
        the second row even though it could, because T3 has low weight
        and thus the information gain would be too low, roughly
        $0.03 < 1$; (2) the sublinear $f$ (log) incentivizes the agent
        to transmit more frequently than the superlinear $f$ (sq)
        does, to maximize its score.}}
  \label{fig:resultsexample}
\end{figure}

\begin{table}[t]
  \centering
  \tabcolsep=0.11cm{
    \resizebox{0.47\columnwidth}{!}{
      \begin{tabular}{c|c|c|c}
        \toprule[1.5pt]
        \textbf{Experiment} & \textbf{Score from Human} & \textbf{\# Info / Timestep} & \textbf{Alg.~\ref{alg:planning} Runtime (sec)}\\
        \midrule[2pt]
        N=4, M=1, f=id & 375 & 0.34 & 6.2\\
        \midrule
        N=4, M=5, f=id & 715 & 0.25 & 6.7\\
        \midrule
        N=6, M=5, f=id & 919 & 0.24 & 24.1\\
        \midrule[1.5pt]
        N=4, M=1, f=sq & 13274 & 0.25 & 4.7\\
        \midrule
        N=4, M=5, f=sq & 33222 & 0.2 & 6.7\\
        \midrule
        N=6, M=5, f=sq & 41575 & 0.19 & 23.6\\
        \midrule[1.5pt]
        N=4, M=1, f=log & 68 & 0.39 & 5.6\\
        \midrule
        N=4, M=5, f=log & 91 & 0.32 & 5.7\\
        \midrule
        N=6, M=5, f=log & 142 & 0.3 & 23.8\\
        \bottomrule[1.5pt]
      \end{tabular}}}
  \quad
  \tabcolsep=0.11cm{
    \resizebox{0.485\columnwidth}{!}{
      \begin{tabular}{c|c|c|c}
        \toprule[1.5pt]
        \textbf{Experiment} & \textbf{Score from Human} & \textbf{\# Info / Timestep} & \textbf{Alg.~\ref{alg:planning} Runtime (sec)}\\
        \midrule[2pt]
        N=5, M=5, f=id & 362 & 0.89 & 0.4\\
        \midrule
        N=5, M=10, f=id & 724 & 1.12 & 2.0\\
        \midrule
        N=10, M=10, f=id & 806 & 1.56 & 48.4\\
        \midrule[1.5pt]
        N=5, M=5, f=sq & 37982 & 0.52 & 0.4\\
        \midrule
        N=5, M=10, f=sq & 99894 & 0.67 & 1.8\\
        \midrule
        N=10, M=10, f=sq & 109207 & 0.71 & 39.7\\
        \midrule[1.5pt]
        N=5, M=5, f=log & 19 & 1.05 & 0.4\\
        \midrule
        N=5, M=10, f=log & 31 & 1.39 & 1.8\\
        \midrule
        N=10, M=10, f=log & 39 & 1.7 & 42.7\\
        \bottomrule[1.5pt]
      \end{tabular}}}
  \caption{\small{Results on the 2D gridworld task (left) and 3D
      continuous task (right) for solving the \mdp\ $\P$ with
      \algref{alg:planning} (no learning; $R_{H}$ is known). Each row
      reports averages over 100 independent trials. $N$ = grid size or
      number of zones, $M$ = number of objects. Planning takes time
      exponential in the environment size. The agent gives information
      less frequently when $f$ is superlinear (sq), and more when $f$
      is sublinear (log).}}
\label{table:results}
\end{table}

\subsection{Domain 1: Search-and-Recover 2D Gridworld Task}
Our first domain is a 2D gridworld in which locations form a discrete
$N \times N$ grid, $M$ objects are scattered across the environment,
and the agent must find and recover all objects. Each object is of a
particular type; the world of object types is known, but all types
need not be present in the environment. The actions that the agent can
perform on each timestep are as follows: \textsc{Move} by one square
in a cardinal direction, with reward -1; \textsc{Detect} whether an
object of a given type is present at the current location, with reward
-5; and \textsc{Recover} the given object type at the current
location, which succeeds with reward -20 if an object of that type is
there, otherwise fails with reward -100.

\begin{figure}[h]
  \centering
    \noindent
    \includegraphics[width=0.32\textwidth]{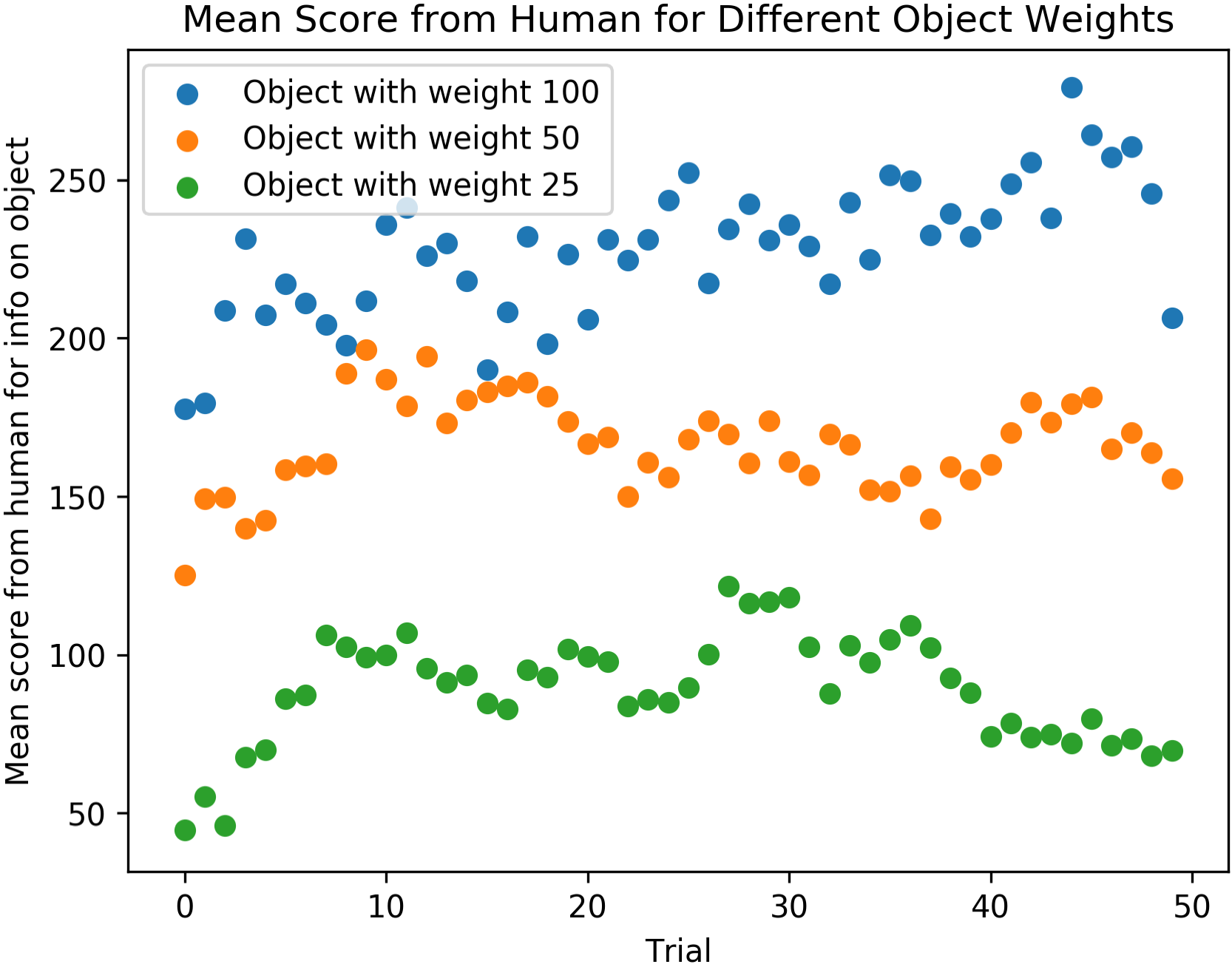}
    \includegraphics[width=0.32\textwidth]{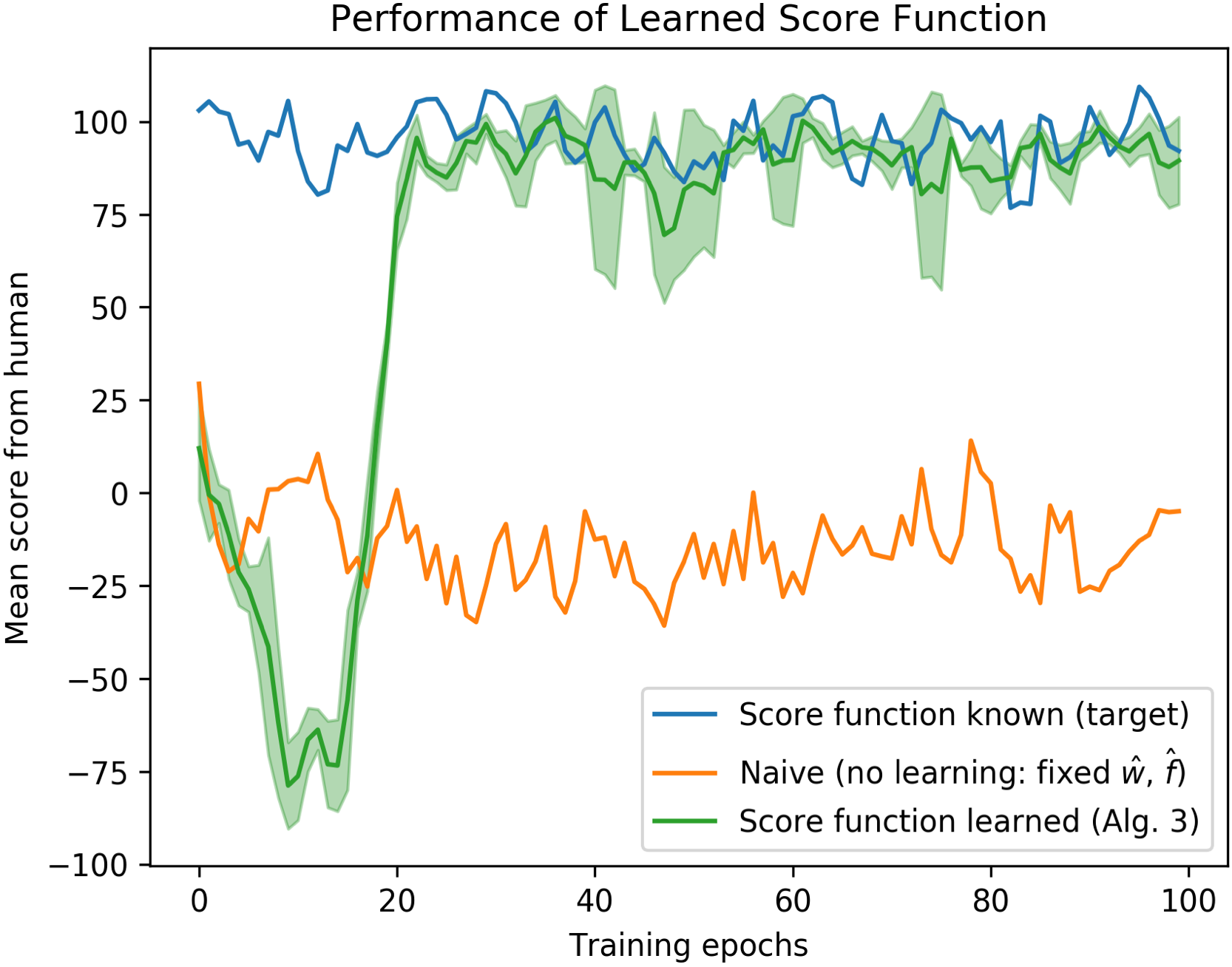}
    \includegraphics[width=0.32\textwidth]{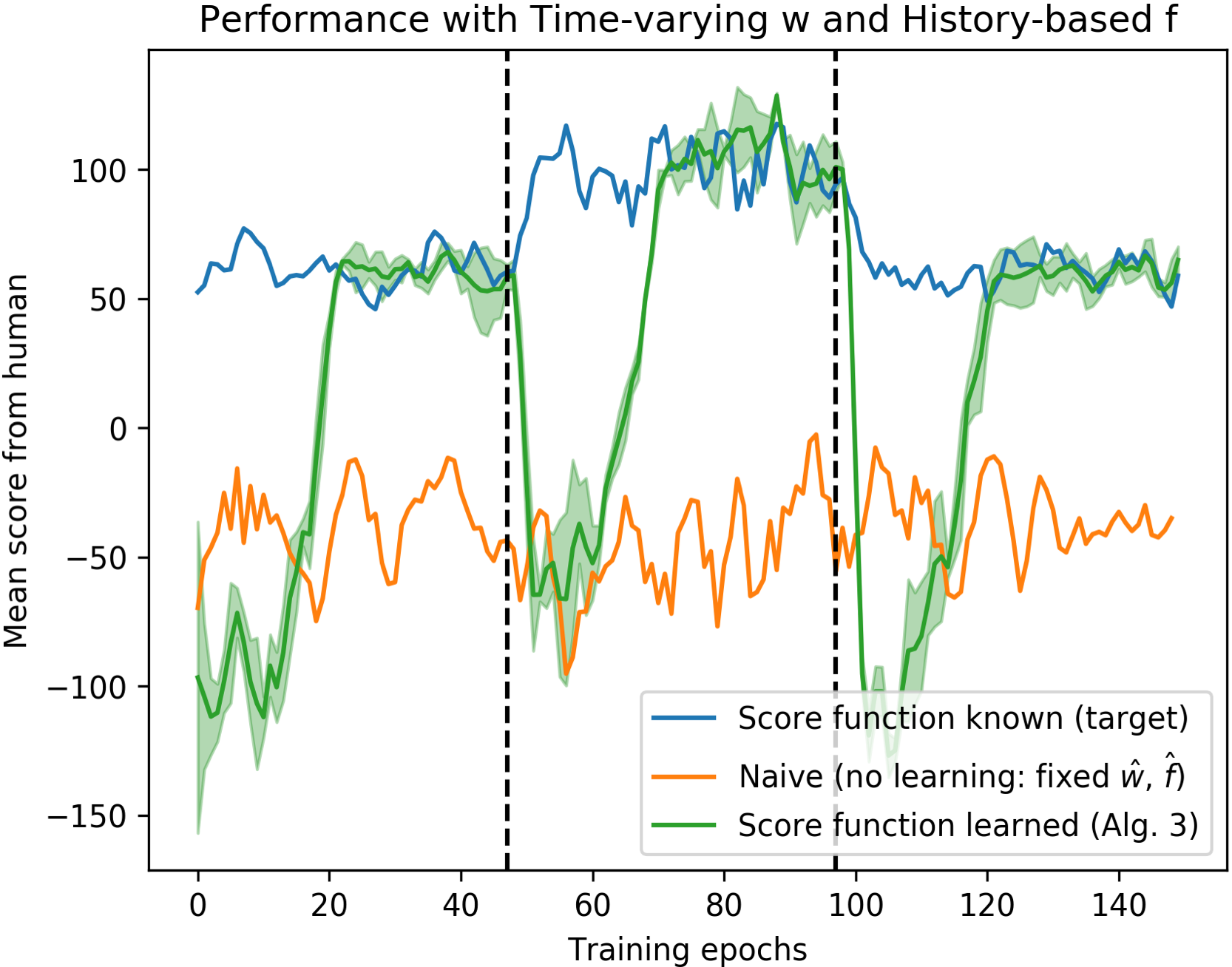}
    \caption{\small{Domain 1 result graphs. \emph{Left.} Confirming
        our intuition, the human gives higher scores for information
        about objects of higher-weighted types. These weights are
        chosen by the human based on their preferences. \emph{Middle.}
        Running \algref{alg:learning}, which learns the true score
        function online, allows the agent to adapt to the human's
        preferences and give good information, earning itself high
        scores. \emph{Right.} We experiment with 1) making $f$
        history-based by penalizing the agent for giving information
        two timesteps in a row, and 2) making $w$ time-varying by
        changing the weights at the training epochs shown by the
        dotted lines. The agent learns to give good information after
        an exploratory period following each change in the human's
        preferences. \emph{Note.} Learning curves are averaged over 5
        independent trials, with standard deviations shaded in
        green.}}
  \label{fig:resultsgw}
\end{figure}

An episode terminates when all $M$ objects have been recovered. To
initialize an episode, we randomly assign each object a type and a
unique grid location. The factored belief representation for both the
agent and the human maps each grid location to a distribution over
what object type (or nothing) is located there, initialized
uniformly. This choice of representation implies that each $w_{i}$ in
the human's weights $w$ represents their interest in receiving
information about object type $i$; for example, the human may
prioritize information regarding valuable objects. The space of
information $\I$ that the agent can select from is: At($t, l$) for
every object type $t$ and location $l$; NotAt($t, l$) for every object
type $t$ and location $l$; and null (no information). Our experiments
vary the grid size $N$, the number of objects $M$, the human's choice
of weights $w$, and the human's choice of $f$. \tabref{table:results},
\figref{fig:resultsexample}, and \figref{fig:resultsgw} show and
discuss our results.

\subsection{Domain 2: Search-and-Recover 3D Continuous Task}
Our second domain is a more realistic 3D robotic environment
implemented in pybullet~\citep{pybullet}. There are $M$ objects in the
world with continuous-valued positions, scattered across N ``zones''
which partition the position space, and the agent must find and
recover all objects. The actions that the agent can perform on each
timestep are as follows: \textsc{Move} to a given pose, with reward
-1; \textsc{Detect} all objects within a cone of visibility in front
of the current pose, with reward -5; and \textsc{Recover} the closest
object within a cone of reachability in front of the current pose,
which succeeds with reward -20 if such an object exists, otherwise
fails with reward -100.

\begin{figure}[h]
  \centering
  \noindent
    \includegraphics[width=0.32\textwidth]{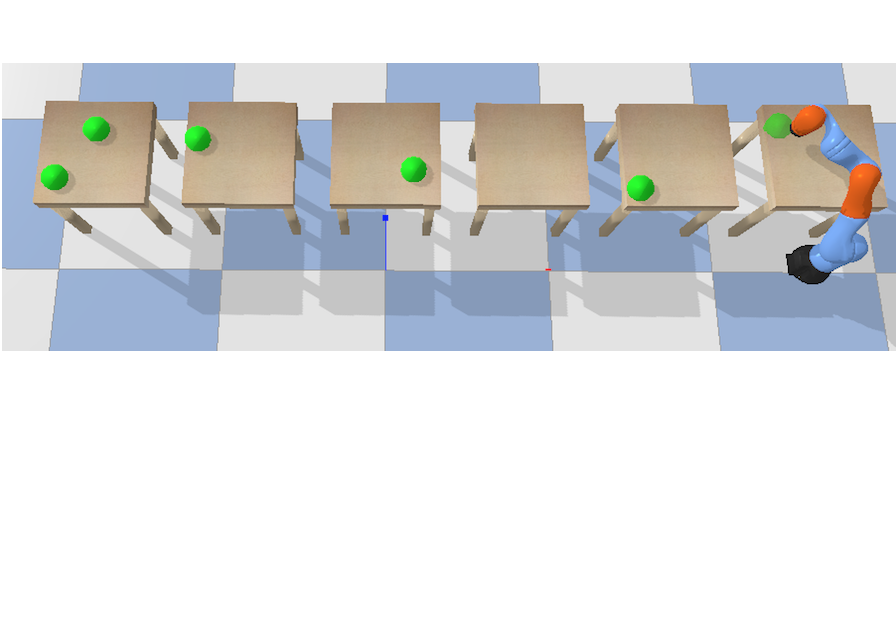}
    \includegraphics[width=0.32\textwidth]{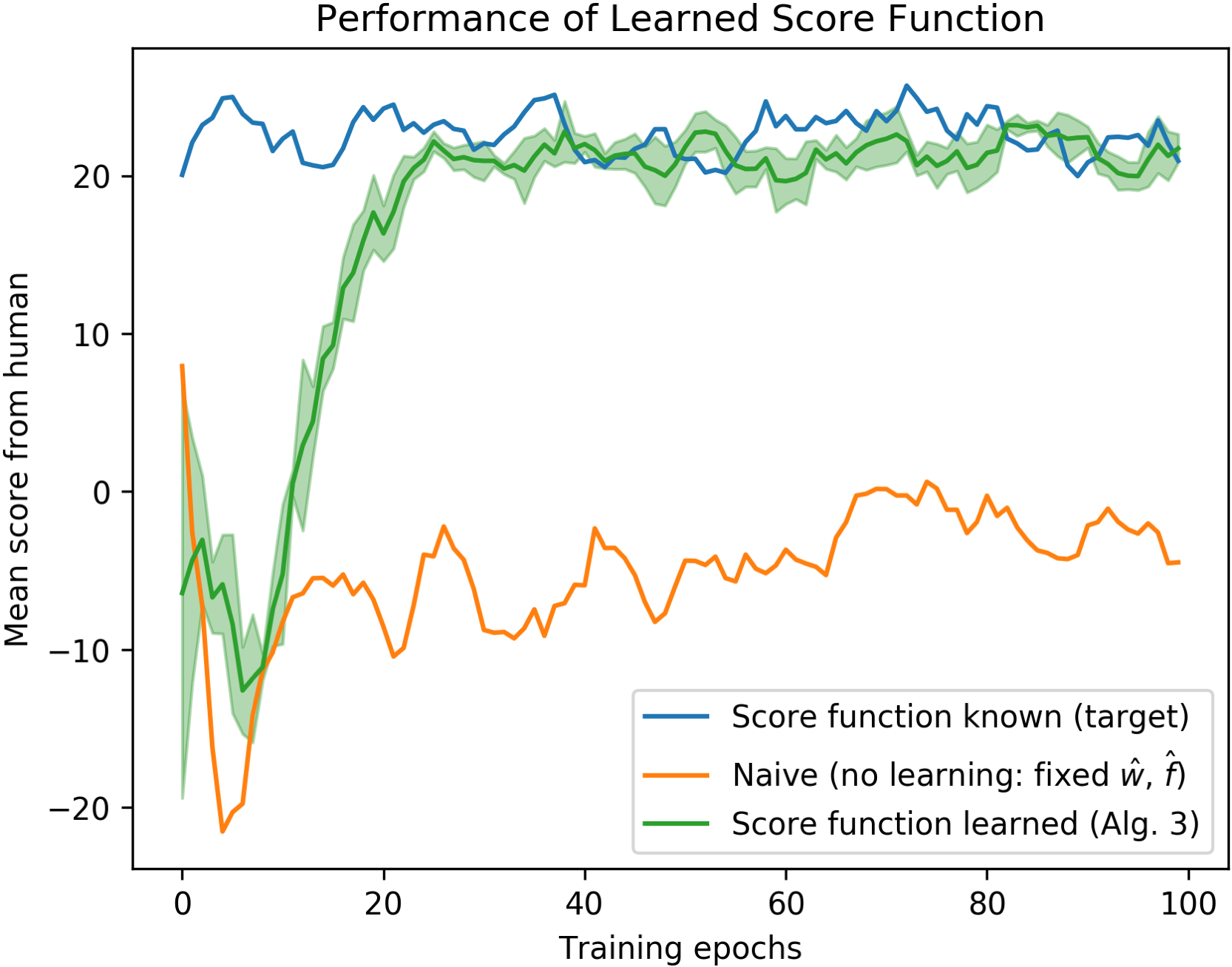}
    \includegraphics[width=0.32\textwidth]{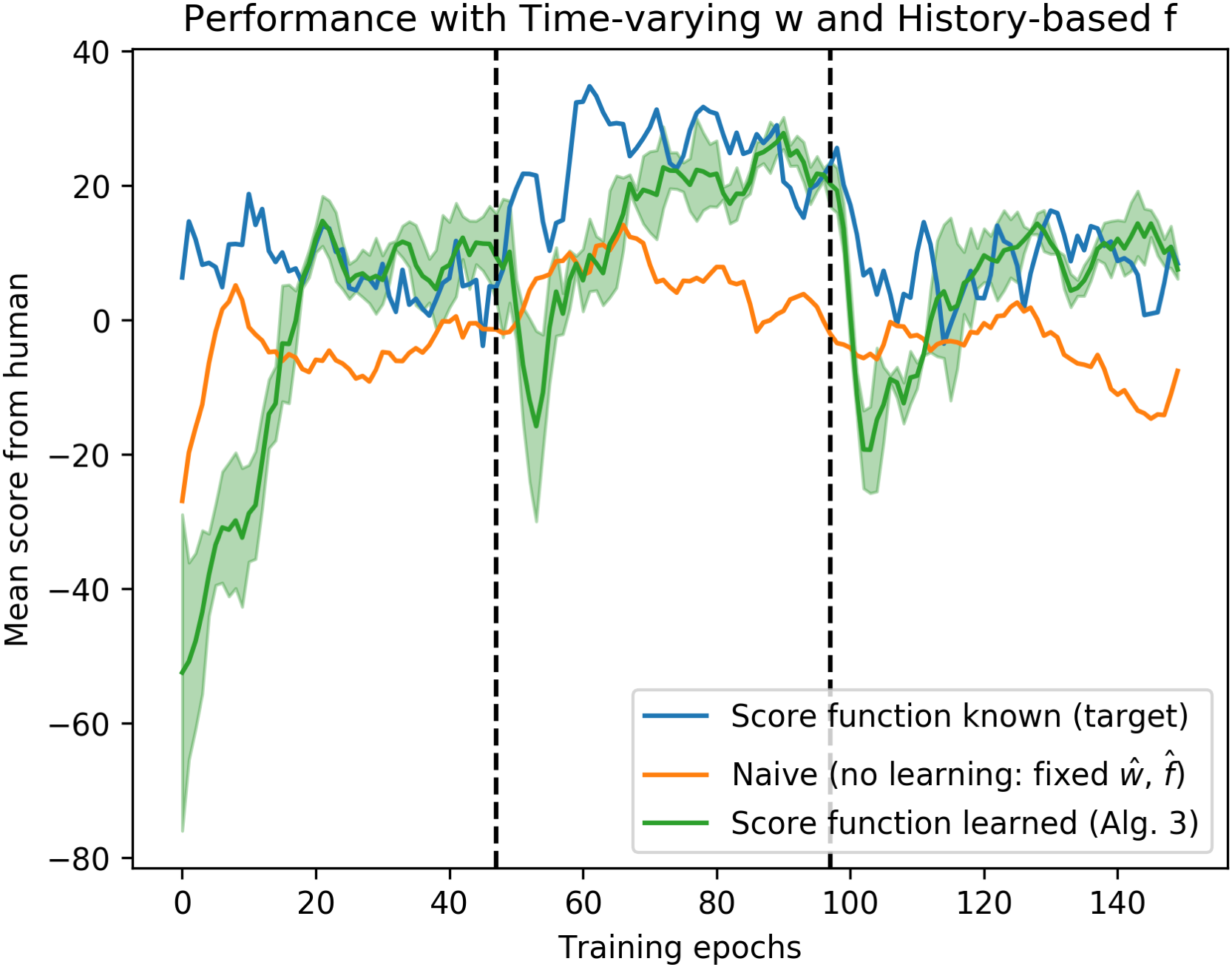}
    \caption{\small{Domain 2 results. \emph{Left.} A pybullet
        rendering of the task. The robot is a blue-and-orange arm, and
        each table is a zone. The green objects are spread across
        table surfaces. \emph{Middle+Right.}  See
        \figref{fig:resultsgw} caption. \emph{Note.} Learning curves
        are averaged over 5 independent trials, with standard
        deviations shaded in green.}}
  \label{fig:resultscont}
\end{figure}

An episode terminates when all $M$ objects have been recovered. To
initialize an episode, we place each object at a random collision-free
position. The factored belief representation for the agent maps each
known object to a distribution over its position, whereas the one for
the human (which must be over a discrete space per our assumptions) maps each known object to a
distribution over which of the $N$ zones it could be in; both are
initialized uniformly. This choice of representation implies that each
$w_{i}$ in the human's weights $w$ represents their interest in
receiving information about zone $i$; for example, the zones could
represent sections of the ocean floor or rooms within a
building on fire. The space of information $\I$ that the agent can select from
is: In($o, z$) for every object $o$ and zone $z$; NotIn($o, z$) for
every object $o$ and zone $z$; and null (no information). Our
experiments vary the number of zones $N$, the number of objects $M$,
the human's choice of weights $w$, and the human's choice of
$f$. \tabref{table:results} and \figref{fig:resultscont} show and
discuss our results.

%% file: conclusion.tex
\section{Conclusion and Future Work}
\label{sec:conclusion}
We have formulated a problem setting in which an agent must act
optimally in a partially observed environment while learning to
transmit information to a human teammate, based on their
preferences. We modeled the human's score as a function of the
weighted information gain of their belief.

One direction for future work is to experiment with settings where the
human has preferences over information about the different
\emph{factors}. Such preferences could be realized by having different
scales of weights across factors, or by calculating the weighted
entropy $S_{w}(B_{H})$ as a weighted sum across factors according to
some other weights $v$ (rather than an unweighted sum as in this
work), possibly learned. Another future direction is to have the agent
learn to generate good candidates for information to transmit, rather
than naively consider all available options in $\I$ at each timestep.